\documentclass[letterpaper]{article} 
\usepackage{aaai25}  
\usepackage{times}  
\usepackage{helvet}  
\usepackage{courier}  
\usepackage[hyphens]{url}  
\usepackage{graphicx} 
\usepackage{natbib}  
\usepackage{caption} 
\usepackage{algorithm}
\usepackage{algorithmic}

\usepackage{newfloat}
\usepackage{listings}
\DeclareCaptionStyle{ruled}{labelfont=normalfont,labelsep=colon,strut=off} 
\floatstyle{ruled}
\newfloat{listing}{tb}{lst}{}
\floatname{listing}{Listing}
\usepackage{tabularx}
\usepackage{multirow}
\usepackage{booktabs}
\usepackage{makecell}
\usepackage{amssymb}
\usepackage[most]{tcolorbox}
\usepackage{hhline}

\newcolumntype{Z}{>{\centering\let\newline\\\arraybackslash\hspace{0pt}}X}
\newcolumntype{P}[1]{>{\centering\arraybackslash}p{#1}}

\title{SVGBuilder: Component-Based Colored SVG Generation with Text-Guided Autoregressive Transformers}
\author{
    Zehao Chen,
    Rong Pan\thanks{Corresponding author.}
}
\affiliations{

    School of Computer Science and Engineering, Sun Yat-sen University \\ 
    Guangzhou, Guangdong, China \\
    chenzh593@mail2.sysu.edu.cn, 
    panr@sysu.edu.cn
}

\begin{document}

\maketitle

\begin{abstract}
  Scalable Vector Graphics (SVG) are essential XML-based formats for versatile graphics, offering resolution independence and scalability. Unlike raster images, SVGs use geometric shapes and support interactivity, animation, and manipulation via CSS and JavaScript. Current SVG generation methods face challenges related to high computational costs and complexity. In contrast, human designers use component-based tools for efficient SVG creation. Inspired by this, SVGBuilder introduces a component-based, autoregressive model for generating high-quality colored SVGs from textual input. It significantly reduces computational overhead and improves efficiency compared to traditional methods. Our model generates SVGs up to 604 times faster than optimization-based approaches. To address the limitations of existing SVG datasets and support our research, we introduce ColorSVG-100K, the first large-scale dataset of colored SVGs, comprising 100,000 graphics. This dataset fills the gap in color information for SVG generation models and enhances diversity in model training. Evaluation against state-of-the-art models demonstrates SVGBuilder's superior performance in practical applications, highlighting its efficiency and quality in generating complex SVG graphics.
\end{abstract}

%
\begin{links}
    \link{Project}{https://svgbuilder.github.io}
\end{links}

\section{Introduction}
\label{sec:introduction}

\begin{figure}[t]
  \centering
  \includegraphics[width=0.9\columnwidth]{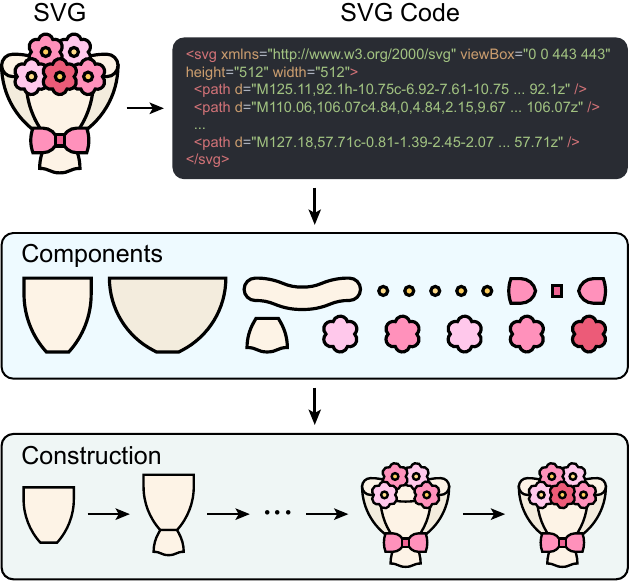}
  \caption{Illustration of component-based SVG generation, with each path in the SVG code considered as a component. The process demonstrates the construction of an SVG from individual components.}
  \label{fig:component_based_svg_generation}
\end{figure}

\begin{figure*}
  \centering
  \includegraphics[width=\textwidth]{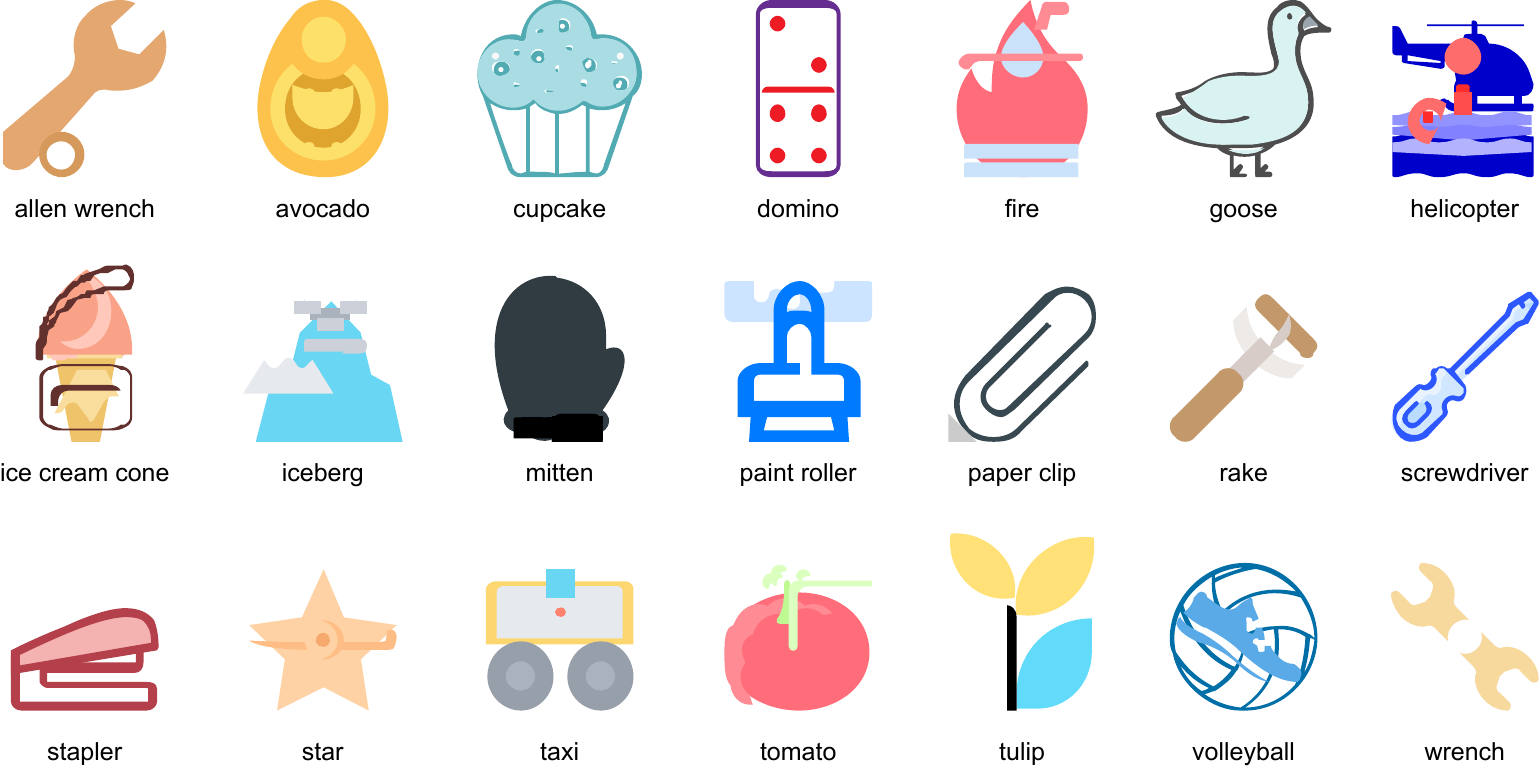}
  \caption{Random samples generated by SVGBuilder.}
  \label{fig:partially_generated_results}
\end{figure*}

Scalable Vector Graphics (SVG) is a widely used XML-based vector image format for defining graphics. Unlike raster graphics, which are composed of a fixed grid of pixels, SVGs are defined by geometric shapes such as points, lines, curves, and polygons, which makes them resolution-independent and infinitely scalable without loss of quality. This inherent scalability offers significant advantages over bitmap images, particularly in web and mobile applications where varying display resolutions and device capabilities are common. Furthermore, SVG supports interactivity and animation, and because it's text-based, it is easily searchable, compressible, and can be manipulated through CSS and JavaScript.

In the current methods for generating SVGs, one of the most straightforward approaches involves using text-to-image generation models \cite{ho2020denoising,rombach2022high} to first create a bitmap, which is then converted into an SVG using vectorization techniques \cite{ma2022towards}. While this method can produce high-quality results, it is computationally intensive and time-intensive. Another approach is optimization-based \cite{jain2023vectorfusion,Xing_2024_CVPR}, starting with a random SVG path and iteratively refining it to match a target SVG. Despite their ability to generate SVGs, these methods have notable drawbacks, such as prolonged processing times and high computational costs \cite{tang2024strokenuwa}. Additionally, some research focuses on unguided random generation of SVG paths \cite{carlier2020deepsvg}, which poses challenges for practical applications due to the lack of directed output. Recently, there has been interest in leveraging autoregressive language models to generate SVGs directly from textual descriptions \cite{wu2023iconshop}. However, SVGs can contain extensive code, which poses a significant challenge for autoregressive language models when generating complex SVGs. The length and complexity of the code increase the difficulty of producing accurate and efficient SVG outputs, often leading to suboptimal performance in intricate design tasks. Additionally, generating SVGs based on individual tokens can easily result in incomplete or invalid paths, further complicating the generation process. Consequently, the creation of sophisticated SVGs using such models remains a significant computational and methodological challenge. For a more detailed discussion on the background of autoregressive language models and SVGs, please refer to supplementary material.

In contrast, human designers often create SVGs using software tools that facilitate component-based design. These tools allow designers to construct images by assembling, scaling, moving, and coloring reusable components. This modular approach simplifies the design process and enhances efficiency, as it enables the reuse of predefined elements and reduces the need to generate complex paths from scratch. Inspired by this observation, our model adopts a component-based strategy for SVG generation. By leveraging predefined graphical components, our approach aims to streamline the SVG creation process, reducing computational overhead and improving the feasibility of generating intricate designs programmatically. As illustrated in Figure~\ref{fig:component_based_svg_generation}, an SVG file consists of SVG code written in XML format, containing tags such as \texttt{<}path\texttt{>}. We consider each \texttt{<}path\texttt{>} as a distinct component. These components are then assembled to construct a complete SVG. This method not only mirrors the intuitive workflows of human designers but also offers a structured and scalable framework for automating SVG generation, thereby addressing some of the limitations inherent in current generation techniques.

Moreover, due to the significant differences between stroke-based SVGs and SVGs with path fills, previous SVG-related datasets primarily focus on stroke-based SVGs. In contrast, datasets containing colored SVGs are mostly stored in bitmap formats or consist of small-scale collections, which are insufficient for training models to learn diverse variations and generation patterns. To address this gap, we develop the first large-scale colored SVG dataset, ColorSVG-100K, aimed at enhancing the capability of models to generate SVGs with rich color information.

To address the aforementioned issues, we present SVGBuilder, a novel model that leverages a component-based, autoregressive approach guided by textual input for colored SVG generation. Our method significantly reduces the generation time and computational expense associated with traditional techniques while ensuring practical applicability by providing meaningful, text-driven guidance for the SVG output. This approach effectively overcomes the challenges of random generation and the difficulties inherent in generating lengthy SVG sequences using autoregressive models. To the best of our knowledge, we are the first to propose a component-based autoregressive model for generating colored SVGs. As shown in Figure \ref{fig:partially_generated_results}, these are some examples of SVGs generated by SVGBuilder.

Compared to state-of-the-art (SOTA) SVG generation models, our approach demonstrates significantly higher efficiency. Specifically, our method generates SVGs up to 604 times faster than optimization-based models currently available. Additionally, our approach outperforms other SOTA models on the FID metric, indicating superior performance. Furthermore, in our case study, it is evident that the graphics produced by our method are of higher quality compared to those generated by other language-based models, showcasing the advantages of our approach in practical applications.

Our contribution lies in three key aspects:

\begin{itemize}
  \item \textbf{Development of SVGBuilder:} We present SVGBuilder, the first component-based autoregressive model for generating colored SVGs guided by textual input. This method significantly reduces the generation time and computational expense associated with traditional techniques, ensuring practical applicability and overcoming the challenges of random generation and lengthy SVG sequences.
  
  \item \textbf{Introduction of ColorSVG-100K Dataset:} We develop the first large-scale dataset of colored SVGs, ColorSVG-100K, aimed at enhancing the capability of models to generate SVGs with rich color information. This dataset addresses the limitations of previous datasets that focused primarily on stroke-based SVGs or small-scale collections.
  
  \item \textbf{Superior Performance and Efficiency:} Our approach not only mirrors the intuitive workflows of human designers by leveraging predefined graphical components but also significantly outperforms SOTA SVG generation models on the FID metric. Our method generates SVGs much faster and produces higher-quality graphics, demonstrating superior efficiency and practical applicability in real-world scenarios.
\end{itemize}

\section{Related Work}
\label{sec:related_work}

\subsection{Text-to-Image Generation}

The field of generating images from text has progressed significantly, evolving through several important phases involving GANs and diffusion models. GANs \cite{goodfellow2014generative} have been instrumental, employing a generator to produce images and a discriminator to evaluate them, refining the output iteratively to align with textual descriptions. While text-conditioned GANs \cite{kang2023scaling, qiao2019mirrorgan, xu2018attngan} have made considerable strides, they often encounter challenges related to scalability and stability when dealing with complex datasets, despite recent advancements \cite{kang2023scaling}. More recently, diffusion models have emerged as the preferred method for text-to-image generation. These models start with Gaussian noise and progressively refine it to generate coherent images. Text-guided diffusion models \cite{nichol2021glide, rombach2022high} incorporate text embeddings either directly or through cross-attention mechanisms to direct the image generation process. Previous research in this domain has predominantly focused on producing raster images with fixed resolutions. In contrast, our work targets the generation of vector icons from text, enabling arbitrary scaling and providing greater flexibility and potential for various applications.

\subsection{SVG Generation}

SVG uses XML code to generate graphics. Currently, a widely used approach is optimization-based, which involves randomly initializing some SVG paths and then optimizing them using a differentiable rasterizer \cite{jain2023vectorfusion, Xing_2024_CVPR}. However, this method is highly time-intensive, as creating an SVG with 24 SVG paths can take over 20 minutes \cite{tang2024strokenuwa}. Some methods generate SVGs purely through random initialization without text guidance \cite{carlier2020deepsvg}. Recent research explores autoregressive methods for generating SVGs \cite{wu2023iconshop}, but the lengthy code of SVG graphics poses challenges for generating complex SVGs using this approach. To address this issue, we adopt a component-based autoregressive generation method for SVGs, significantly improving the generation speed and overcoming the difficulties of using autoregressive techniques for complex SVGs, while achieving high-quality results.

\section{Methodology}
\label{sec:methodology}

\begin{figure*}[t]
  \centering
  \includegraphics[width=\textwidth]{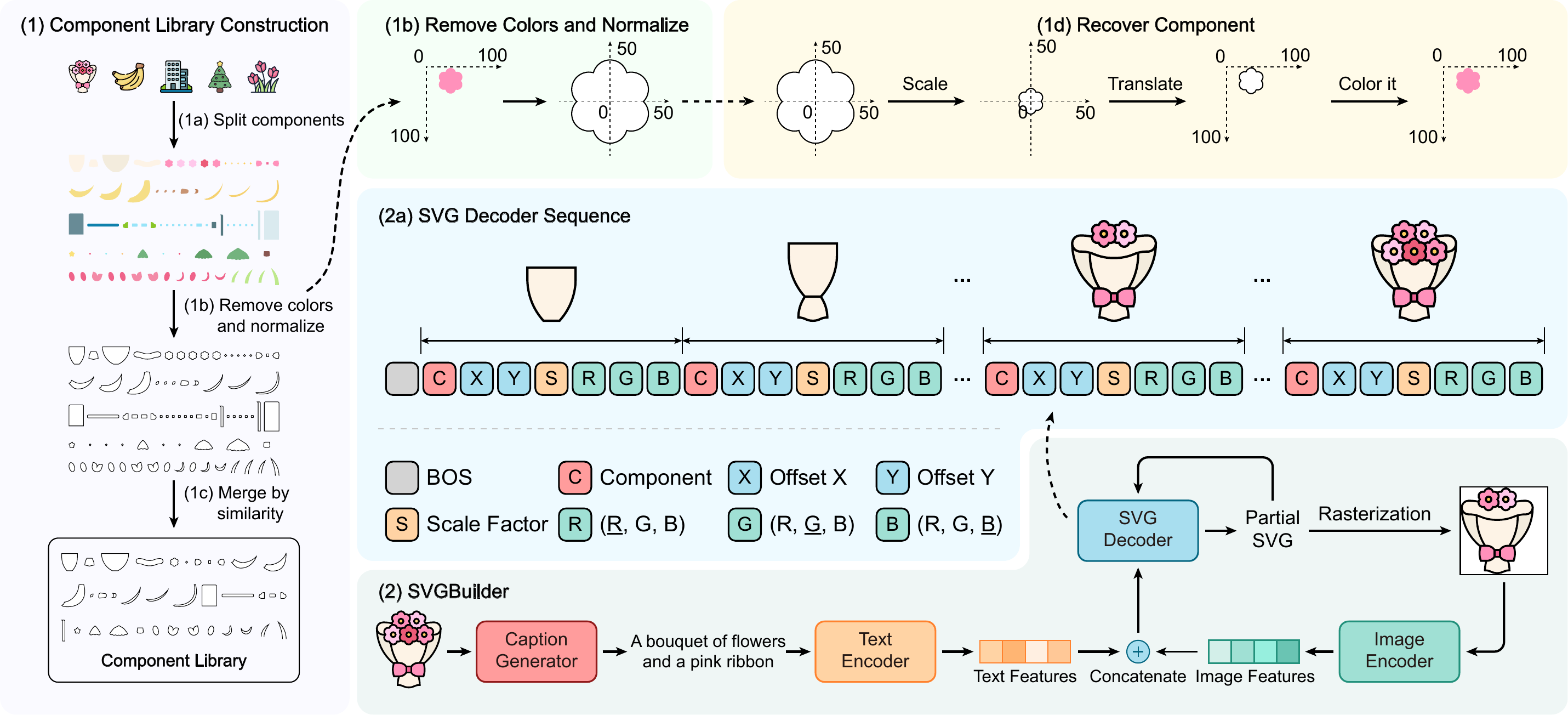}
  \caption{System framework overview. The framework comprises two primary steps. The first step is Component Library Construction, which includes the sub-process of Remove Colors and Normalize. The second step involves the SVGBuilder's operations, which encompass the SVG Decoder Sequence and the Recover Component processes.}
  \label{fig:model}
\end{figure*}

In this section, we provide a comprehensive explanation of our proposed method for SVG generation. Our system framework consists of two primary parts. The first part involves the construction of the component library, which serves as the foundation for the SVG generation process. The second part is our SVG generation model, which utilizes the established component library to produce high-quality SVGs. The overall system framework, highlighting the interaction between these parts, is illustrated in Figure \ref{fig:model}.

\subsection{Component Library Construction}

In our system, the generation of models is component-based, making the establishment of a component library the critical first step, as illustrated in Figure \ref{fig:model} (1). In Figure \ref{fig:model} (1a), we begin by decomposing all SVGs in the training set into individual paths, with each path forming a unique component. These isolated components exhibit a wide variety of appearances, necessitating a thorough reorganization.

Next, as shown in Figure \ref{fig:model} (1b), we remove the colors from these components and normalize them. Initially, the geometric centers of these components do not align with the origin, complicating the scaling and normalization process. To address this issue, we calculate the geometric center of each component and translate it to the origin. We then scale the component such that its longest dimension measures 100 units, ensuring that the extents along both the positive and negative axes are 50 units. This process reorganizes and normalizes the components, making them ready for subsequent use in the model.

The initial component library contains many duplicate components. For example, flower shapes in a bouquet may be identical except for their colors. Therefore, it is necessary to merge these redundant components. As depicted in Figure \ref{fig:model} (1c), we start by merging paths that are identical in the SVGs. This initial merging reduces the number of components but does not eliminate all redundancies. To further refine the component library, we convert all components into $100 \times 100$ pixel images and then transform these images into grayscale. In these grayscale images, the areas enclosed by the paths are represented by 1 (white), and the background is represented by 0 (black). We then compute the Jaccard similarity index between each pair of images, sorting the similarity scores in descending order. Using a union-find data structure, we set a merging threshold and examine the similarity scores sequentially. If the similarity between two components meets or exceeds the threshold, we merge them in the union-find structure. The root nodes of these merged groups are considered the final components in the library. Subsequently, we replace occurrences of the child nodes with their corresponding root nodes in the training set. This process results in a refined and consolidated component library, ensuring minimal redundancy and optimal component organization for the model.

When the normalized components need to be restored, as shown in Figure \ref{fig:model} (1d), it is necessary to place them in the correct positions, similar to how a human would arrange elements when drawing in software. This involves scaling the components according to a scale factor and translating them based on offset values on the X and Y axes (Offset X and Offset Y). This ensures accurate placement of the components. Finally, the generated RGB tokens are used to colorize the component, restoring its original appearance.

\subsection{SVGBuilder}

Our SVG generation model comprises three main components: the Text Encoder, the Image Encoder, and the SVG Decoder, as illustrated in Figure \ref{fig:model} (2). During training, the original dataset typically provides minimal text prompts, usually limited to category labels like ``flowers'' without further description of the SVG. To address this limitation, we introduce a Caption Generator, which is employed solely during training and not during inference.

Initially, we rasterize the SVG into an image and feed this image into the Caption Generator, which produces a caption, such as ``A bouquet of flowers and a pink ribbon.'' This textual description is then input into the Text Encoder to generate text features. At the same time, we generate image features by inputting a blank image into the Image Encoder. The Image Encoder outputs image features, which are concatenated with the text features. These combined features are then input into the SVG Decoder. The SVG Decoder also takes an SVG Decoder Sequence as input to generate the next token. The partially generated SVG is rasterized into an image, which is subsequently fed back into the Image Encoder. This process repeats until an end token is generated or the maximum sequence length is reached. The role of the Image Encoder in this process is to allow the model to understand semantics not only from textual or path-based perspectives but also from visual perspectives. By incorporating the Image Encoder, the model can recognize and comprehend the components it selects at each step, including their layout and color. This dual learning mechanism ensures that the model grasps the specific visual attributes and spatial arrangements of the elements, enhancing its overall semantic understanding and performance in tasks involving both textual and visual data.

Specifically, as shown in Figure \ref{fig:model} (2a), the SVG Decoder Sequence consists of several tokens. Each complete component is represented by seven tokens. The sequence begins with a BOS (beginning of sequence) token, followed by a token representing the component. The next tokens indicate the translation offsets (Offset X and Offset Y) and the scaling factor (Scale Factor), specifying how the component should be placed. Finally, the sequence includes tokens for color (R, G, and B tokens), representing the RGB values. This sequence is repeated for each component until the entire SVG is generated.

\section{Experiments}
\label{sec:experiments}

In this section, we present a series of experiments conducted to substantiate the efficacy of our proposed SVGBuilder methodology. This section elucidates our evaluation metrics, describes our baseline comparisons, and offers a comprehensive analysis of the results derived from our experimental investigations. We also introduce our proposed ColorSVG-100K dataset, which serves as a critical resource for our experiments. Additionally, we conducted SVG complexity analysis, ablation experiments, and a case study to further validate our approach.

\subsection{Dataset}

The creation of the ColorSVG-100K dataset arises from the significant limitations observed in existing SVG datasets. Available SVG datasets are predominantly either black and white or stored as raster images, lacking the essential vector properties necessary for many applications. Such limitations pose substantial challenges for tasks requiring high-quality, scalable, and color-rich SVGs. The primary motivation behind developing the ColorSVG-100K dataset is to address these gaps and provide a robust foundation for research and applications in the field of colored SVG generation. This dataset, comprising 100,000 richly colored SVGs, is the first of its kind on a large scale. It not only fills the void of color information in existing datasets but also leverages the advantages of the SVG format, such as scalability and editability. For more information about the ColorSVG-100K dataset, including the motivation for its creation, the specific construction process, dataset statistics, and examples, please refer to the supplementary materials.

\subsection{Experimental Settings}

\paragraph{Evaluation Metrics}

We adopt evaluation metrics from SkexGen, IconShop \cite{wu2023iconshop}, and StrokeNUWA. Specifically, we use the Fréchet Inception Distance (FID) \cite{heusel2017gans} to quantify the distance between the image features of the generated and ground-truth SVGs. Additionally, we utilize two types of CLIPScore \cite{pmlr-v139-radford21a}: CLIPScore-T2I, which measures the similarity between the rasterized images of the generated SVGs and the textual prompts used for generation, and CLIPScore-I2I, which measures the similarity between the rasterized images of the generated SVGs and the rasterized images of the ground-truth SVGs. Furthermore, we assess the ``Uniqueness'' and ``Novelty'' of the generated SVGs, which are derived from SkexGen. ``Uniqueness'' indicates the proportion of generated data occurring only once among all generated results, while ``Novelty'' refers to the proportion of generated data absent from the training set. We also incorporate the Human Preference Score (HPS) \cite{wu2023human} to evaluate user satisfaction with the generated outputs. These metrics collectively provide a comprehensive evaluation of the quality, originality, and performance of our SVG generation methodology. Lastly, we measure the generation speed per SVG to assess the efficiency of our method.

\paragraph{Baselines}

In this paper, we compare our method with various SOTA open-source models. VectorFusion \cite{jain2023vectorfusion} utilizes diffusion models trained on pixel representations of images to generate SVGs, without requiring large datasets of captioned SVGs. By optimizing a differentiable vector graphics rasterizer \cite{li2020differentiable}, VectorFusion extracts semantic knowledge from a pretrained diffusion model, resulting in high-quality vector graphics across various styles. CLIPDraw \cite{frans2022clipdraw} employs a pretrained CLIP language-image encoder to synthesize novel drawings by optimizing the similarity between the generated drawing and the given description. It operates over vector strokes rather than pixel images, producing simple, human-recognizable shapes. DiffSketcher \cite{xing2024diffsketcher}, based on a pretrained diffusion model, optimizes a set of Bézier curves to create vectorized free-hand sketches that retain the structure and visual details of the subject. We re-implemented IconShop using Flan-T5 \cite{chung2024scaling} as the backbone, employing autoregressive transformers to sequentialize and tokenize SVG paths and textual descriptions, resulting in improved icon synthesis capabilities. Additionally, we explored the capabilities of GPT-3.5 and GPT-4 \cite{achiam2023gpt} in SVG generation, demonstrating advancements in large language models (LLMs) for this task. All these methods are text-guided, allowing us to evaluate the performance and effectiveness of the proposed SVGBuilder methodology against the current SOTA methods in the field of SVG generation.

\paragraph{Implementation Details}

In our experiments, when rasterizing SVGs, we place them on a white background. For constructing the ColorSVG-100K dataset, ``Category Calibration" corrects misclassified SVGs. Both the Text Encoder and Image Encoder in this stage, as well as during model training and FID calculation, are CLIP models \cite{pmlr-v139-radford21a}. During SVG processing, we utilize the DeepSVG Library. The Caption Generator is BLIP-2 \cite{li2023blip}, and the SVG Decoder is based on GPT-2 \cite{radford2019language}. The batch size for training is set to 16, with a learning rate of $6 \times 10^{-4}$. We use the AdamW optimizer for 50 epochs of training. The experiments are conducted on a single NVIDIA A800 GPU. For more implementation details, please refer to the supplementary materials\ref{sec:more_implementation_details}.

\subsection{Performance Comparison}

\begin{table*}[]
  \centering
  \begin{tabularx}{\textwidth}{ccccccc>{\centering\arraybackslash}m{1.3cm}c}
  \toprule
  \multirow{2}{*}{Model Type}                         & \multirow{2}{*}{Model}        & \multirow{2}{*}{FID ($\downarrow$)}   & \multicolumn{2}{c}{CLIPScore}                                               & \multirow{2}{*}{HPS ($\uparrow$)} & \multirow{2}{*}{Uniqueness ($\uparrow$)} & \multirow{2}{*}{Novelty ($\uparrow$)} & \multirow{2}{*}{\makecell{Generation Speed \\ per SVG ($\downarrow$)}}        \\ \cmidrule(lr){4-5}
                                                      &                               &                                       & \multicolumn{1}{l}{T2I ($\uparrow$)} & \multicolumn{1}{l}{I2I ($\uparrow$)} &                                                                                                                                                                                                      \\
  \midrule
  \multirow{3}{*}{\makecell{Optimization \\ Based}}   & VectorFusion                  & 20.62                                 & 27.86                                & \underline{77.84}                    & \textbf{\underline{17.65}}        & \textbf{\underline{100}}                 & \textbf{\underline{100}}              & $\approx$ 13.69 min ($1.0 \times$)                                \\
                                                      & DiffSketcher                  & \underline{20.28}                     & 27.04                                & 77.70                                & 17.62                             & \textbf{\underline{100}}                 & \textbf{\underline{100}}              & $\approx$ 4.27 min ($3.2 \times$)                                 \\
                                                      & CLIPDraw                      & 30.11                                 & \textbf{\underline{28.46}}           & 73.13                                & 17.51                             & \textbf{\underline{100}}                 & \textbf{\underline{100}}              & $\approx$ \underline{2.92} min ($\underline{4.7} \times$)         \\
  \midrule
  \multirow{4}{*}{\makecell{Language \\ Based}}       & GPT-3.5                       & 33.19                                 & 22.05                                & 77.19                                & 16.96                             & 96.2                                     & \textbf{\underline{100}}              & 10.8 s ($76.1 \times$)                                            \\
                                                      & GPT-4                         & 31.85                                 & 22.57                                & 77.88                                & 16.99                             & 98.4                                     & \textbf{\underline{100}}              & 23.8 s ($34.5 \times$)                                            \\
                                                      & IconShop                      & 41.03                                 & 21.96                                & 73.53                                & 16.57                             & 97.8                                     & \textbf{\underline{100}}              & 8.52 s ($96.4 \times$)                                            \\
                                                      & Ours                          & \textbf{\underline{15.93}}            & \underline{22.76}                    & \textbf{\underline{79.28}}           & \underline{17.10}                 & \underline{99.2}                         & \textbf{\underline{100}}              & \textbf{\underline{1.36}} s ($\mathbf{\underline{604.0}} \times$) \\
  \bottomrule
  \end{tabularx}
  \caption{Performance comparison between SVGBuilder and the baseline models on the ColorSVG-100K dataset. \underline{Underlined} values indicate the best performance within each model type, while \textbf{bold} values indicate the best overall performance across all models.}
  \label{tab:performance_comparison}
\end{table*}

We evaluate the performance of different models on the ColorSVG-100K dataset, as illustrated in Table \ref{tab:performance_comparison}. The models are categorized into two types: optimization-based and language-based. The arrows in the table indicate the preferred direction, with down arrows indicating lower values are better, and up arrows indicating higher values are better.

In the Table \ref{tab:performance_comparison}, CLIPScore is divided into T2I, which represents the CLIPScore between the generated SVG image and the text description, and I2I, which represents the CLIPScore between the generated SVG image and the real SVG image. From Table \ref{tab:performance_comparison}, it is evident that the optimization-based models have lower FID scores, indicating that the images they generate are closer to the distribution of real images in the test set. In contrast, the language-based models, except for ours, generally have higher FID scores, suggesting that previous methods fail to accurately fit the dataset's distribution.

Regarding CLIPScore, optimization-based methods generally achieve higher T2I scores. This might be because these models generate images that align better with the distribution of images on which the CLIP model is originally trained. The CLIP model might not have been extensively trained on SVG datasets, making it less effective at capturing the flat 2D nature of SVG graphics in CLIPScore-T2I scores. In contrast, the CLIPScore-I2I measures the similarity between generated and real SVG images, and hence is not affected by this limitation. Notably, our model also achieves the highest CLIPScore among the language-based models.

Similarly, since HPS uses the same CLIP model, it tends to favor SVG generation models that optimize from bitmap images. Regarding Uniqueness, the random nature of optimization-based models results in a 100\% score, while language-based models exhibit slightly lower uniqueness due to potential similar outputs from similar inputs. Our model, however, achieves the highest uniqueness score among the language-based models. All models achieve a 100\% score in Novelty.

In terms of Generation speed per SVG, the values in parentheses indicate the speed-up factor relative to VectorFusion. It is clear that optimization-based models take a significantly longer time to generate an SVG, with VectorFusion taking approximately 13.69 minutes and even the fastest, CLIPDraw, taking 2.92 minutes. In contrast, language-based models take only a few seconds, drastically reducing generation time. Our model achieves a generation time of 1.36 seconds, representing a $604.0 \times$ speed-up compared to VectorFusion, which underscores the potential and advantages of language-based models in SVG generation.

\subsection{SVG Complexity Analysis}

\begin{table}[]
  \centering
  \begin{tabularx}{\columnwidth}{c|cccc}
  \toprule
  \multirow{2}{*}{Model} & \multicolumn{4}{c}{SVG Complexity}                                                                                     \\ \cmidrule(r){2-5}
                         & 0-25\%                      & 25-50\%                        & 50-75\%                    & 75-100\%                   \\
  \midrule
  VectorFusion           & 32.70                       & 30.26                          & \underline{29.82}          & 31.34                      \\
  DiffSketcher           & \underline{29.09}           & \underline{29.24}              & 30.62                      & \underline{30.96}          \\
  CLIPDraw               & 42.86                       & 40.43                          & 38.48                      & 39.37                      \\
  \midrule
  GPT-3.5                & 32.18                       & 38.59                          & 42.52                      & 47.00                      \\
  GPT-4                  & 32.28                       & 37.62                          & 40.46                      & 43.88                      \\
  IconShop               & 40.80                       & 45.34                          & 50.00                      & 53.60                      \\
  Ours                   & \textbf{\underline{20.47}}  & \textbf{\underline{23.36}}     & \textbf{\underline{26.93}} & \textbf{\underline{30.55}} \\
  \bottomrule
  \end{tabularx}
  \caption{Comparison of FID scores across different SVG complexity levels for various models. \underline{Underlined} values indicate the best performance within each model type, while \textbf{bold} values indicate the best overall performance across all models.}
  \label{tab:complexity_analysis}
\end{table}

We further analyze the impact of varying SVG complexities on model performance. We sort the SVGs in ascending order based on the number of paths and divide them into four levels of complexity: 0-25\%, 25-50\%, 50-75\%, and 75-100\%. The 0-25\% range represents the simplest SVGs with the fewest paths, indicating the lowest complexity, while the 75-100\% range encompasses the SVGs with the most paths, indicating the highest complexity. We evaluate the performance of different models across these complexity levels using the FID metric, as shown in Table \ref{tab:complexity_analysis}. Our model achieves the best results among all models across all levels of complexity. It is observed that as the complexity of SVGs increases, the performance of optimization-based models does not deteriorate, whereas the performance of language-based models worsens. This may be because the randomly initialized bitmaps are generally more complex than SVGs.

\subsection{Ablation Study}

\begin{table}[]
  \centering
  \begin{tabularx}{\columnwidth}{lcccc}
  \toprule
  \multirow{2}{*}{Model}    & \multirow{2}{*}{FID ($\downarrow$)} & \multicolumn{2}{c}{CLIPScore}                                               \\ \cmidrule(lr){3-4}
                            &                                     & \multicolumn{1}{l}{T2I ($\uparrow$)} & \multicolumn{1}{l}{I2I ($\uparrow$)} \\
  \midrule
  Ours                      & \textbf{15.93}                      & \textbf{22.76}                       & \textbf{79.28}                       \\
  \midrule
  w/o Caption Generator     & 17.40                               & 21.94                                & 78.23                                \\
  w/o Image Encoder         & 16.90                               & 22.58                                & 77.97                                \\
  w/o training Text Encoder & 21.75                               & 21.73                                & 78.04                                \\
  w/o SVG filling           & 17.39                               & 22.64                                & 77.79                                \\
  \bottomrule
  \end{tabularx}
  \caption{Performance comparison of SVGBuilder without different components.}
  \label{tab:ablation_study}
\end{table}

To evaluate the impact of different components on the overall performance of our model, we conduct an ablation study. The components subjected to ablation include the Caption Generator, Image Encoder, whether the Text Encoder is trained, and whether color fills are used in SVG during inference. We test each component sequentially, and the results are presented in Table \ref{tab:ablation_study}. In these experiments, we observe both the FID and CLIPScore metrics. Since changes in the CLIPScore are not significant, we primarily focus on the FID metric. The results indicate that removing the Image Encoder results in the least performance loss on the FID metric, suggesting that the model does not effectively utilize the information from the Image Encoder. Removing the Caption Generator has a moderate impact on FID, as using the original SVG categories as input text lacks the data augmentation benefits provided by diverse captions. Freezing the Text Encoder leads to a significant decline in FID performance, highlighting the importance of training the Text Encoder for maintaining model performance. Lastly, not filling colors during inference and only performing outlines also results in a decrease in FID.

\subsection{Case Study}

\begin{figure}
  \centering
  \includegraphics[width=\columnwidth]{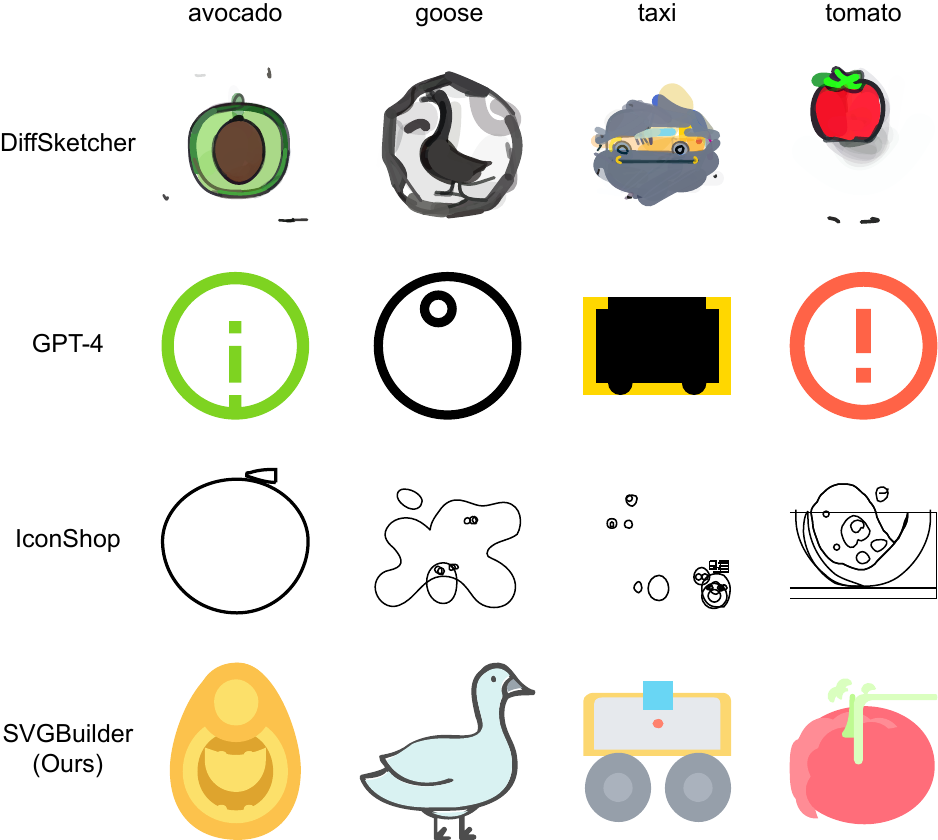}
  \caption{Examples of SVGs generated by different models. DiffSketcher represents the optimization-based model, while the others are language-based models.}
  \label{fig:case_study}
\end{figure}

In our case study, we select SVGs generated by different models for comparative analysis. We focus on language-based models, including GPT-4, IconShop, and our model. For comparison, we also include DiffSketcher, an optimization-based model with the lowest FID score. The results are illustrated in Figure \ref{fig:case_study}.

DiffSketcher produces visually impressive results, accurately capturing the original shapes and colors of objects, likely due to the application of the Diffusion model. However, some artifacts in the form of odd lines appear during the bitmap-to-SVG optimization process. The output from GPT-4 aligns well with the objects in terms of color and somewhat meets the contour requirements, but without accompanying text, the contours are hard to interpret. IconShop generates accurate outlines except for the taxi example, which performs poorly. This discrepancy might be due to the greater complexity of SVGs in our dataset compared to those used by IconShop, indicating its suitability for simpler SVG generation tasks.

Our model's outputs match the objects in both color and contour. However, the ``avocado'' example is influenced by the seed, affecting the overall hue, and the ``taxi'' appears abstract. Overall, our model approaches the expression of the optimization-based models, demonstrating its potential.

\section{Limitations}
\label{sec:limitations}

While SVGBuilder demonstrates strong performance in component-based autoregressive SVG generation, it does have certain limitations. Our ColorSVG-100K dataset fills a notable gap in the realm of colored SVG datasets, yet each path within the dataset currently lacks explicit semantics. This limitation forces the model to implicitly infer these semantics, which may not always be reliable. Additionally, the construction of the component library is crucial within this framework. We normalize components through translation and scaling based on similarity for merging. However, incorporating additional geometric transformations, such as rotation and reflection, could potentially further reduce the size of the component library. Although the similarity-based merging process is relatively time-consuming, it is conducted offline, representing a one-time computational cost. Furthermore, when utilizing a union-find data structure for merging, selecting the parent node is critical. Even with a similarity threshold in place, nodes within the same set may have varying degrees of similarity to the parent node due to the non-transitive nature of similarity. Additionally, the component-based approach can be more challenging to innovate compared to stroke-based methods. In our language model training, we use cross-entropy as the loss function, treating numbers as separate tokens. However, cross-entropy may not effectively capture geometric relationships. Looking ahead, we aim to address these challenges and make incremental advancements in the field of component-based autoregressive SVG generation.

\section{Conclusion}
\label{sec:conclusion}

In this study, we present SVGBuilder, an innovative component-based autoregressive model designed for the generation of colored SVGs, alongside the introduction of the ColorSVG-100K dataset. Through rigorous evaluation, our approach demonstrates SOTA performance, surpassing both optimization-based and language-based models in the FID metric. Remarkably, SVGBuilder not only excels in visual quality but also offers a substantial $604 \times$ enhancement in generation speed compared to optimization-based models, showing great potential for future applications.

\section*{Acknowledgements}
\label{sec:acknowledgements}

The authors are supported by the National Key Research and Development Program of China (Grant No. 2020YFA0712500), the National Natural Science Foundation of China (Grant No. 12126609), and the Pazhou Lab (Huangpu) Research and Development Project, funded under Grant No. 2023K0601. The authors would also like to thank National Supercomputer Center in Guangzhou for providing high performance computational resources.

\bibliography{chen}

\appendix
\clearpage

\section{Background}
\label{sec:background}

\subsection{Autoregressive Language Models}

Autoregressive language models predict the next token in a sequence based on previous tokens. Formally, given a sequence of tokens \( a = (a_1, a_2, \ldots, a_T) \), the probability of the sequence is decomposed as:
\begin{equation}
  P(a) = \prod_{t=1}^{T} P(a_t \mid a_{1:t-1}).
\end{equation}
Here, \( P(a_t \mid a_{1:t-1}) \) represents the probability of the token \( a_t \) given the preceding tokens \( a_{1:t-1} \). This approach allows the model to generate coherent text by sampling each token sequentially. The training objective is to maximize the likelihood of the observed sequence, which is achieved by minimizing the negative log-likelihood:
\begin{equation}
  \mathcal{L} = - \sum_{t=1}^{T} \log P(a_t \mid a_{1:t-1}).
\end{equation}

Autoregressive models, such as GPT \cite{brown2020language}, have shown significant success in natural language processing tasks, leveraging large-scale pre-training on diverse text corpora to generate fluent and contextually appropriate text.

In this paper, we utilize autoregressive language models for text-guided, component-based SVG generation.

\subsection{Scalable Vector Graphic (SVG)}

Scalable Vector Graphics (SVG) is an XML-based markup language for describing two-dimensional vector graphics. Unlike raster images, SVGs are mathematically defined, allowing them to be scaled to any size without loss of quality, ensuring sharp images on all devices. Additionally, SVGs are typically smaller, improving web page load times. These features make SVG a versatile and powerful tool for creating high-quality, scalable, and interactive graphics.

\paragraph{SVG Representation}

In SVGs, various tags are employed to represent specific shapes, such as \texttt{<}rect\texttt{>} for rectangles, \texttt{<}circle\texttt{>} for circles, and \texttt{<}path\texttt{>} for paths. This variety in tag usage leads to inconsistency in data representation and hinders component-level manipulation and scaling. To address this issue, we standardize all shapes to \texttt{<}path\texttt{>} tags, following the approach outlined in DeepSVG \cite{carlier2020deepsvg}. Within the \texttt{<}path\texttt{>} tag, we use uniform commands for drawing, including $\mathrm{M}$ (Move to), $\mathrm{L}$ (Line to), $\mathrm{C}$ (Cubic Bézier), and $\mathrm{Z}$ (Close path), as shown in Table \ref{tab:svg_command}. Other basic shapes can be represented using a sequence of Bézier curves and lines. Following StrokeNUWA \cite{tang2024strokenuwa}, a simplified SVG $\mathcal{G} = \{ \mathcal{P}_i \}_{i=1}^N$ is represented by $N$ SVG paths, where each path $\mathcal{P}_i$ consists of $M_i$ basic commands: $\mathcal{P}_i = \{ \mathcal{C}_i^j \}_{j=1}^{M_i}$. Here, $\mathcal{C}_i^j$ denotes the $j$-th command in the $i$-th path. Each basic command $\mathcal{C} = (T, \mathcal{V})$ includes a command type $ T \in \{ \mathrm{M}, \mathrm{L}, \mathrm{C} \} $ and its corresponding positional argument $\mathcal{V}$.

\begin{table}[t]
  \centering
  \begin{tabularx}{\columnwidth}{cccc}
    \toprule
    Name                       & Symbol       & Arguments                                                               & Visualization \\
    \midrule
    Move To                    & $\mathrm{M}$ & $(x_1, y_1), (x_2, y_2)$                                                & \makecell{\includegraphics[width=1.5cm]{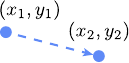}} \\
    Line To                    & $\mathrm{L}$ & $(x_1, y_1), (x_2, y_2)$                                                & \makecell{\includegraphics[width=1.5cm]{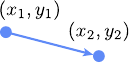}} \\
    \makecell{Cubic \\ Bézier} & $\mathrm{C}$ & \makecell{$(x_1, y_1), (x_2, y_2)$ \\ $(q_1^x, q_1^y), (q_2^x, q_2^y)$} & \makecell{\includegraphics[width=2cm]{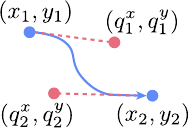}} \\
    \makecell{Close \\ Path}   & $\mathrm{Z}$ & $\varnothing$                                                           & \makecell{\includegraphics[width=2cm]{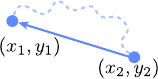}} \\
    \bottomrule
  \end{tabularx}
  \caption{List of simplified SVG commands with their names, symbols, arguments, and visualizations. The $(x_1, y_1)$ typically represents the end-position of the previous command.}
  \label{tab:svg_command}
\end{table}

\section{The ColorSVG-100K Dataset}
\label{sec:colorsvg_100k_dataset}

\subsection{Dataset Motivation}

ColorSVG-100K may play a crucial role in advancing various research areas and practical applications. For instance, it supports the development and training of sophisticated algorithms for generating and manipulating colored SVGs, enhancing the capabilities of graphic design software, and improving data visualization techniques. Moreover, it facilitates better user experiences in web and mobile applications by providing high-quality, scalable vector graphics that are visually appealing.

The significance of ColorSVG-100K lies in its potential to drive innovation and research in numerous fields, including computer graphics, machine learning, and digital art. By providing a comprehensive dataset filled with diverse and detailed SVGs, we aim to empower researchers and developers to explore new frontiers in vector graphic technology and achieve breakthroughs that are previously constrained by the lack of appropriate data.

\subsection{Dataset Construction}

\begin{figure*}[t]
  \centering
  \includegraphics[width=\textwidth]{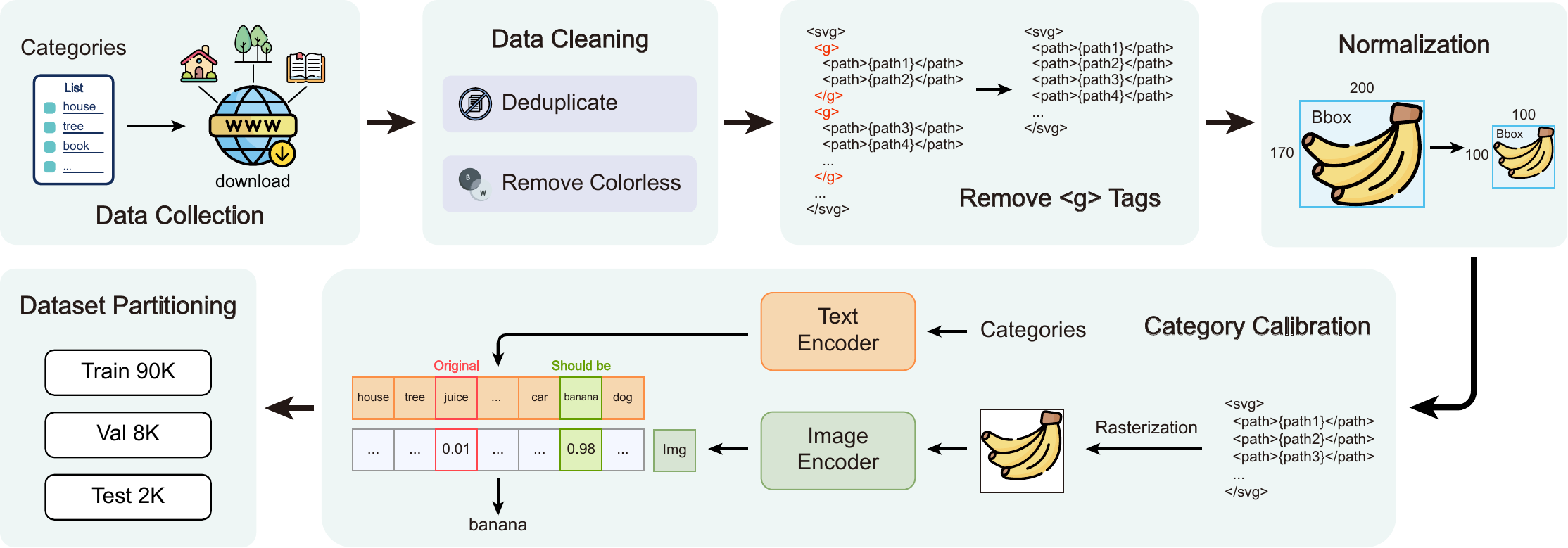}
  \caption{The construction process of the ColorSVG-100K dataset, including the steps of data collection, data cleaning, removal of \texttt{<}g\texttt{>} tags, normalization, category calibration, and dataset partitioning.}
  \label{fig:build_dataset_process}
\end{figure*}

The construction process of the ColorSVG-100K dataset, illustrated in Figure \ref{fig:build_dataset_process}, involves a meticulous and multi-step approach designed to ensure the highest quality and relevance of the data. This subsection details each stage of the dataset's creation, from initial data collection to final dataset partitioning, highlighting the methodologies and considerations involved in each step.

\paragraph{Data Collection}
The first step in constructing the ColorSVG-100K dataset is data collection. This involves sourcing a wide range of SVGs from various online repositories and open-source platforms, aiming to gather SVGs with different styles, themes, and levels of complexity. Ensuring diversity is crucial to create a comprehensive dataset capable of supporting various research and application needs. Based on the categories from the Icon645 dataset \cite{lu2021iconqa}, further filtering and additions are made to identify common categories. Subsequently, SVGs are collected and downloaded from the web according to this refined category list. The SVGs collected at this stage are quite raw and of varying quality.

\begin{figure}[h]
  \centering
  \includegraphics[width=\columnwidth]{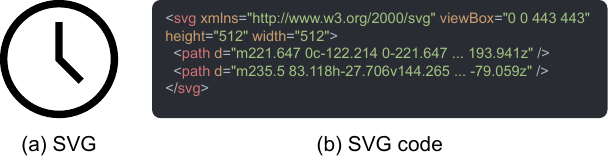}
  \caption{Colorless SVG diagram. (a) represents an SVG graphic, (b) shows the corresponding SVG code for (a).}
  \label{fig:colorless_svg_diagram}
\end{figure}

\paragraph{Data Cleaning}
Following data collection, the dataset undergoes a rigorous cleaning process. This step is essential for removing any corrupted, incomplete, or low-quality SVGs that could negatively impact algorithm performance. The initial data often contains duplicates and colorless SVGs, which contradict the core purpose of our dataset. Therefore, we remove these duplicates and colorless SVGs, along with any corrupted or incomplete SVGs, to ensure the dataset's reliability for subsequent processing and analysis. In the process of building the dataset, we perform the following steps to remove colorless SVGs from the originally collected set. First, we examine the SVGs to determine if they lack a ``fill'' attribute. If an SVG does not contain the ``fill'' attribute, it is preliminarily considered to be a black-only SVG, as illustrated in Figure \ref{fig:colorless_svg_diagram}. Such files are rendered in default black and are subsequently removed. Here, we disregard other potential fill methods. Next, we rasterize the SVGs to convert them into bitmap images and check if the image colors are exclusively black (0) and white (255). If this condition is met, the SVG is deleted. Additionally, we eliminate SVGs where black is the predominant color. To do this, we assess the ratio of black pixels to other colored pixels (excluding the white background). If the black pixels outnumber the other colored pixels, the corresponding SVG is also removed. This thorough cleaning is crucial to enhance the overall quality and consistency of the dataset.

\paragraph{Remove \texttt{<}g\texttt{>} Tags}
The next step is the removal of \texttt{<}g\texttt{>} tags from the SVGs. The \texttt{<}g\texttt{>} tags in SVGs are used to group multiple elements, which can complicate the processing and manipulation of individual graphic components. However, these \texttt{<}g\texttt{>} tags often carry attributes that affect the \texttt{<}path\texttt{>} elements within them. Therefore, instead of simply deleting the \texttt{<}g\texttt{>} tags, we transfer their attributes to the child nodes before removing the tags. We also utilize the svglib from the DeepSVG \footnote{\url{https://github.com/alexandre01/deepsvg}}, enhancing it further to remove parts containing the \texttt{<}g\texttt{>} tag in the original SVGs, resulting in a flattened combination of paths. By doing so, we simplify the structure of the SVGs, making them more accessible and easier to handle for various computational tasks, while preserving the integrity of the graphics.

\paragraph{Normalization}
Normalization is a crucial step to ensure uniformity across the dataset. Bounding box (Bbox) is used to define the spatial extent of graphics within SVGs, providing crucial information for positioning and scaling in various applications. However, the Bboxes in the collected SVGs are inconsistent, with varying sizes and positions that could complicate subsequent processing and analysis. To address this, we standardize all Bboxes to a uniform size of $100 \times 100$ units. This involves adjusting the dimensions and positions of the SVGs so that they fit within the standardized Bbox. By normalizing the Bboxes, we ensure that all SVGs adhere to a common spatial framework, making them more reliable and consistent for use in different research and application contexts. This process facilitates easier handling and manipulation of the graphics, enabling more accurate algorithm training.

\paragraph{Category Calibration}
Subsequently, we organize and categorize each SVG in the dataset, tallying the number of SVGs in each category. We sort these categories in descending order and select the top 500 categories with the highest counts to form the dataset's categories. As a result, our dataset contains 500 distinct SVG categories. To enhance the usability of the dataset, we perform category calibration on the SVGs. Initially, the SVGs collected based on a category list might be misclassified, with SVGs not matching their assigned categories. Therefore, reclassification is necessary to ensure accurate categorization according to the actual content of the SVGs. First, we rasterize the SVGs to obtain images. Then, we use the CLIP \cite{pmlr-v139-radford21a} model to assist in the reclassification process. The rasterized images are fed into the CLIP Image Encoder, while the category labels are fed into the CLIP Text Encoder. We compute the similarity between the resulting features, and the highest-scoring match is assigned as the output label, determining the correct category for each SVG. At this stage, the total number of SVGs still exceeds 100K. We then use the CLIP model to classify all rasterized SVGs based on these 500 categories. If an SVG has a low confidence score in the model's classification compared to its original category, we remove it from the dataset, refining the total to 100K. Subsequently, we use the same method to correct misclassified SVGs with high confidence scores.

\paragraph{Dataset Partitioning}
The final step in constructing the dataset is partitioning it into training, validation, and test sets. Considering that SVG or image generation models may require substantial time for generation, we allocate 8K for the validation set and 2K for the test set. To maintain a balanced distribution of categories and visual features, we use stratified sampling to divide the original 100K dataset. This ensures that each subset accurately represents the diversity of the full dataset. As a result, the training set comprises 90K SVGs, the validation set includes 8K SVGs, and the test set contains 2K SVGs.

Through these detailed and systematic steps, we have constructed the ColorSVG-100K dataset to be a robust and valuable resource for the research community, offering a rich collection of colored SVGs that meet high standards of quality and diversity.

\subsection{Dataset Statistics}

\begin{figure}[h]
  \centering
  \includegraphics[width=\columnwidth]{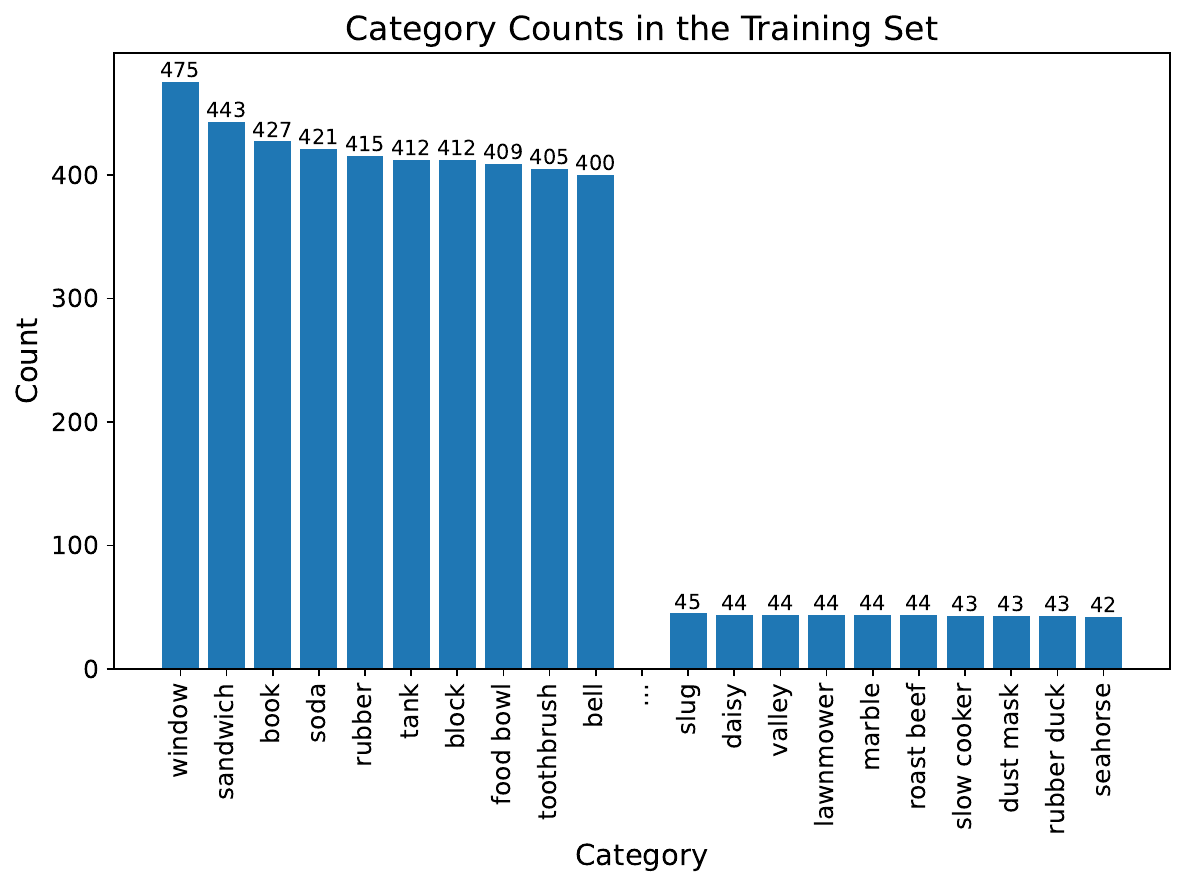}
  \caption{Category counts in the training set.}
  \label{fig:category_counts_in_the_training_set}
\end{figure}

We conduct a detailed statistical analysis of the ColorSVG-100K dataset. In the training set, we categorize SVG samples according to their respective classes, tallying the number of samples in each category. The results are then sorted in descending order, with intermediate results omitted for clarity, as illustrated in Figure \ref{fig:category_counts_in_the_training_set}. The category with the highest number of samples contains up to 475 instances, while the category with the fewest samples has around 40 instances. This imbalance in the dataset arises from the varying prevalence of different SVG categories available online, where more common categories have more samples compared to the rarer ones.

\begin{figure}[h]
  \centering
  \includegraphics[width=\columnwidth]{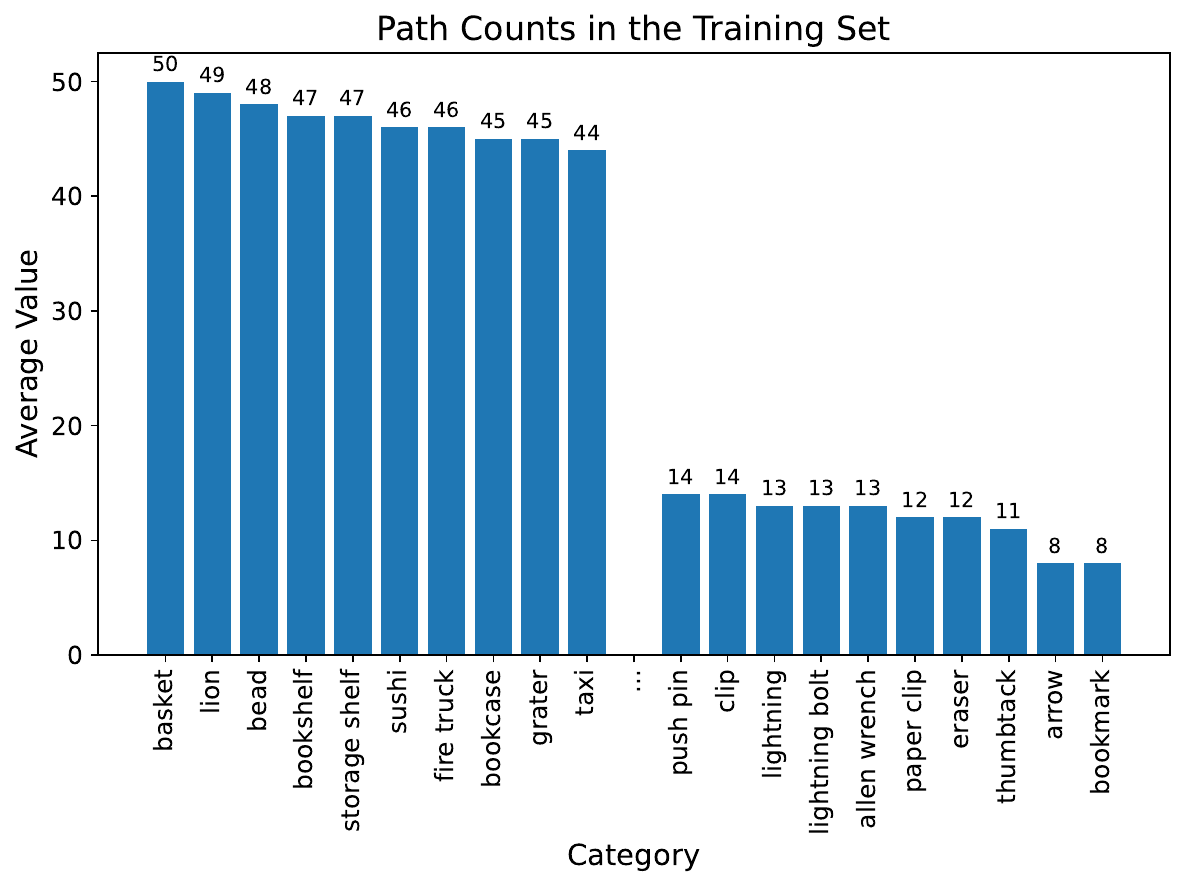}
  \caption{Average path counts across different categories in the training set.}
  \label{fig:path_counts_in_the_training_set}
\end{figure}

We also analyze the average number of paths per category in the training set to assess the complexity of different categories. This analysis, sorted in descending order and with intermediate results omitted for clarity, is presented in Figure \ref{fig:path_counts_in_the_training_set}. The category with the highest average number of paths is ``basket'' followed by ``lion'' indicating these categories have more intricate designs with numerous lines, thus higher complexity. In contrast, the categories with the fewest average paths are ``arrow'' and ``bookmark'' suggesting these SVGs are less complex.

\begin{table}[]
  \centering
  \begin{tabularx}{\columnwidth}{cccZ}
  \toprule
  Dataset                         & \makecell{Dimension \\ (per Category)} & Type    & Value \\
  \midrule
  \multirow{6}{*}{Training Set}   & \multirow{3}{*}{Samples}               & Maximum & 475   \\
                                  &                                        & Minimum & 42    \\
                                  &                                        & Average & 180   \\ \cmidrule(r){2-4}
                                  & \multirow{3}{*}{Paths}                 & Maximum & 1620  \\
                                  &                                        & Minimum & 2     \\
                                  &                                        & Average & 26    \\
  \midrule
  \multirow{6}{*}{Validation Set} & \multirow{3}{*}{Samples}               & Maximum & 42    \\
                                  &                                        & Minimum & 4     \\
                                  &                                        & Average & 16    \\ \cmidrule(r){2-4}
                                  & \multirow{3}{*}{Paths}                 & Maximum & 1058  \\
                                  &                                        & Minimum & 2     \\
                                  &                                        & Average & 26    \\
  \midrule
  \multirow{6}{*}{Test Set}       & \multirow{3}{*}{Samples}               & Maximum & 11    \\
                                  &                                        & Minimum & 1     \\
                                  &                                        & Average & 4     \\ \cmidrule(r){2-4}
                                  & \multirow{3}{*}{Paths}                 & Maximum & 680   \\
                                  &                                        & Minimum & 2     \\
                                  &                                        & Average & 26    \\
  \bottomrule
  \end{tabularx}
  \caption{Statistical information of the dataset, including maximum, minimum, and average values across different dimensions for various subsets.}
  \label{tab:dataset_statistics}
\end{table}

We further summarize the maximum, minimum, and average values across different dimensions for various subsets of the dataset, as shown in Table \ref{tab:dataset_statistics}.

\subsection{Dataset Examples}

\begin{figure*}[h]
  \centering
  \includegraphics[width=\textwidth]{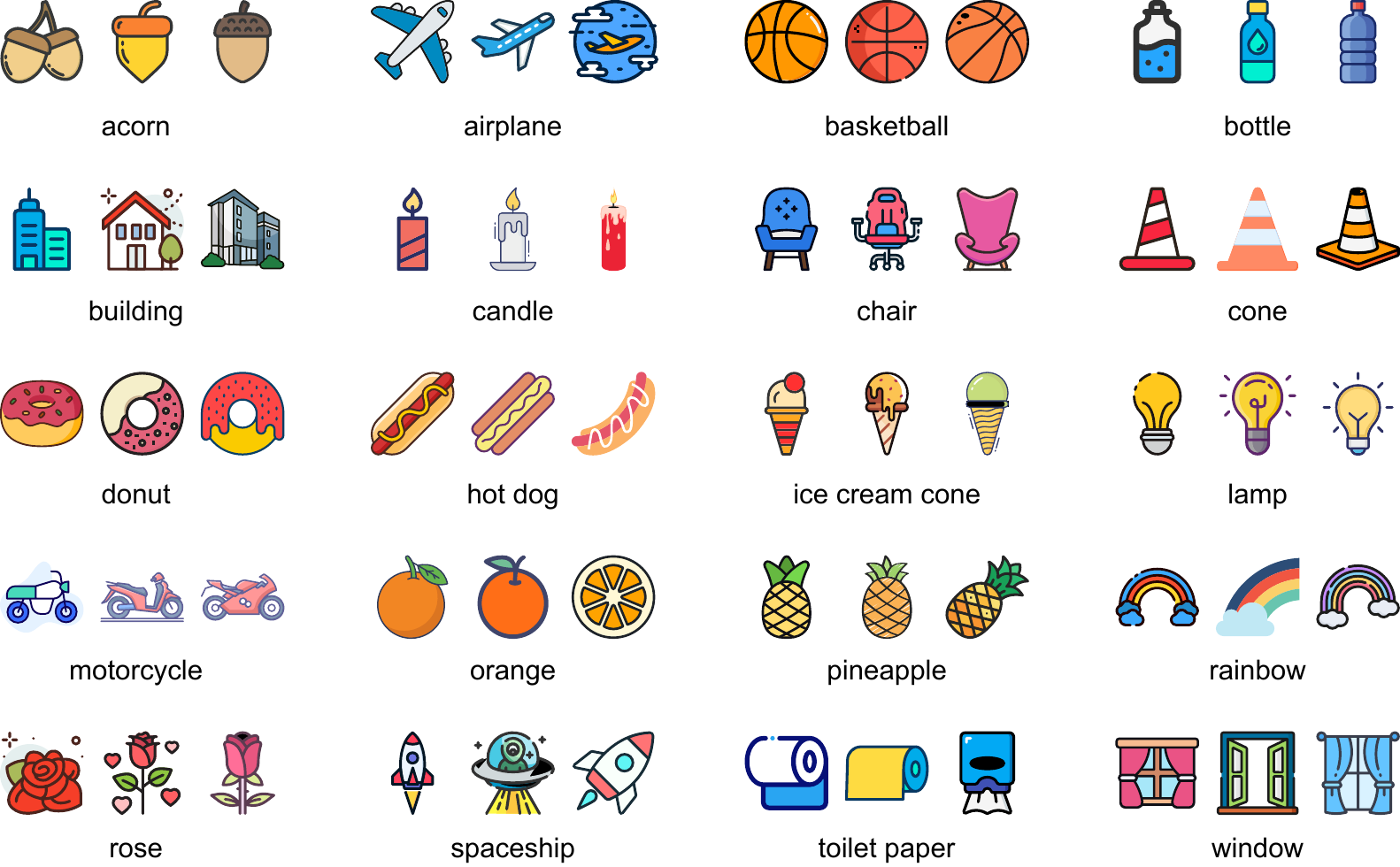}
  \caption{Randomly selected examples from various categories in the training set.}
  \label{fig:dataset_examples}
\end{figure*}

We randomly select several categories from the training set, showcasing three randomly chosen examples from each category, as illustrated in Figure \ref{fig:dataset_examples}.

\section{More Implementation Details}
\label{sec:more_implementation_details}

The baseline models employed in our study, including VectorFusion, CLIPDraw, and DiffSketcher, are available through PyTorch-SVGRender \footnote{\url{https://github.com/ximinng/PyTorch-SVGRender}}. We use these models with their original parameter settings, making necessary adjustments as required. The hyperparameters that differ from the original settings appear in Table \ref{tab:hyperparameter_baselines}. In the component merging process based on similarity, we set the similarity threshold to 0.92. To re-implement IconShop, we train the model on the ColorSVG-100K dataset using a single NVIDIA A800 GPU with hyperparameters of 50 epochs, batch size of 50, learning rate of \(6 \times 10^{-4}\), and model dimension of 768. The specific version of GPT-3.5 is referred to as ``gpt-3.5-turbo,'' while the specific version of GPT-4 is called ``gpt-4-turbo.''

\begin{table}
  \centering
  \begin{tabularx}{\columnwidth}{ZcZ}
    \toprule
    Model                         & Hyperparameter          & Value \\
    \midrule
    \multirow{5}{*}{VectorFusion} & num\_iter               & 300   \\
                                  & num\_paths              & 64    \\
                                  & path\_reinit/stop\_step & 600   \\
                                  & K                       & 1     \\
                                  & sds/num\_iter           & 800   \\
    \midrule
    \multirow{2}{*}{CLIPDraw}     & image\_size             & 100   \\
                                  & num\_paths              & 64    \\
    \midrule
    \multirow{6}{*}{DiffSketcher} & image\_size             & 100   \\
                                  & num\_iter               & 1000  \\
                                  & num\_paths              & 64    \\
                                  & optim\_opacity          & False \\
                                  & optim\_width            & True  \\
                                  & optim\_rgba             & True  \\
    \bottomrule
  \end{tabularx}
  \caption{Hyperparameters differing from the original settings for baselines models.}
  \label{tab:hyperparameter_baselines}
\end{table}

When performing inference with these models, each model receives the same input prompt, where \texttt{<}category\texttt{>} specifies the SVG category:

\begin{tcolorbox}[colback=yellow!10!white, colframe=yellow!75!black]
  \texttt{<}category\texttt{>}. minimal flat 2d vector icon. lineal color. trending on artstation.
\end{tcolorbox}

Given that GPT-3.5 and GPT-4 generate SVG code directly from the input prompt, unlike the aforementioned baseline models, it is necessary to modify the prompt accordingly. The template for the prompt is as follows:

\begin{tcolorbox}[colback=yellow!10!white, colframe=yellow!75!black]
  Please generate an SVG code for a minimal flat 2D vector icon with lineal color style based on the following keyword: \texttt{<}category\texttt{>}. Ensure the design is simple and adheres to a flat design aesthetic. Do not include any additional text or information.
\end{tcolorbox}

In the process of generating SVG code directly with GPT-3.5 and GPT-4, the models may not always produce the code in a single attempt. Instead, they might provide a series of steps or sometimes fail to output any SVG code at all. To address this, we use regular expressions to extract the SVG code from the output. If the initial attempt does not yield SVG code, we repeat the process multiple times until the regular expression successfully identifies and extracts the SVG code.

To address the time-intensive nature of optimization-based methods, we follow StrokeNUWA's setup, randomly selecting 500 samples as the test set for fair and efficient evaluation.

\end{document}